# GRATIS: GeneRAting TIme Series with diverse and controllable characteristics

Yanfei Kang,* Rob J Hyndman,† and Feng Li‡

**Abstract**

The explosion of time series data in recent years has brought a flourish of new time series analysis methods, for forecasting, clustering, classification and other tasks. The evaluation of these new methods requires either collecting or simulating a diverse set of time series benchmarking data to enable reliable comparisons against alternative approaches. We propose GeneRAting TIme Series with diverse and controllable characteristics, named GRATIS, with the use of mixture autoregressive (MAR) models. We simulate sets of time series using MAR models and investigate the diversity and coverage of the generated time series in a time series feature space. By tuning the parameters of the MAR models, GRATIS is also able to efficiently generate new time series with controllable features. In general, as a costless surrogate to the traditional data collection approach, GRATIS can be used as an evaluation tool for tasks such as time series forecasting and classification. We illustrate the usefulness of our time series generation process through a time series forecasting application.

Keywords: Time series features; Time series generation; Mixture autoregressive models; Time series forecasting; Simulation.

# 1 Introduction

With the widespread collection of time series data via scanners, monitors and other automated data collection devices, there has been an explosion of time series analysis methods developed in the past decade or two. Paradoxically, the large datasets are often also relatively homogeneous in the industry domain, which limits their use for evaluation of general time series analysis methods (Keogh and Kasetty, 2003; Muñoz et al., 2018; Kang et al., 2017). The performance of any time series mining algorithm depends on the diversity of the training data, so that the evaluation of the algorithm can be generalized to a wide range of future data (Smith-Miles and Bowly, 2015).

*School of Economics and Management, Beihang University, Beijing 100191, China. Email: yanfeikang@buaa.edu.cn, ORCID: https://orcid.org/0000-0001-8769-6650.

†Department of Econometrics and Business Statistics, Monash University, Clayton, Victoria, 3800, Australia. Email: Rob.Hyndman@monash.edu, ORCID: https://orcid.org/0000-0002-2140-5352.

‡School of Statistics and Mathematics, Central University of Finance and Economics, 100081 Beijing, China. Email: feng.li@cufe.edu.cn, ORCID: https://orcid.org/0000-0002-4248-9778, Corresponding author.

As Keogh and Kasetty (2003) argue, after extensive analysis on highly diverse datasets, there is "*a need for more comprehensive time series benchmarks and more careful evaluation in the data mining community*". Although some attempts have been made to alleviate these issues in certain time series tasks, such as the widely used UCR archive (Dau et al., 2018) for time series classification (see e.g., Ismail Fawaz et al. (2019)), and the M4 dataset for the most recent time series forecasting competition (Makridakis et al., 2018), the time series area lacks diverse and controllable benchmarking data for algorithm evaluation, compared to the popularly used benchmarking datasets in other data mining domains like ImageNet (Deng et al., 2009) for imaging analysis and UCI machine learning repository (Dua and Karra Taniskidou, 2017) for general machine learning algorithms evaluation.

In recent years, research shows that it is also possible to use generated data for algorithm learning under certain application domains, which give great potential for exploring simulated data. One such example is the well-known "Alpha Zero" (Silver et al., 2017), being able to learn from simulated games based on self-play without human input for guidance. Simulated datasets are also extremely useful when a researcher needs good control over the generated data and a careful design of the data generator in order to test the generalizability and weaknesses of their methods. Such examples can be found in evaluating and comparing statistical methods for early temporal detection of outbreaks based on simulated data (Lotze et al., 2010; Bédubourg and Le Strat, 2017), examining neural network forecasting through a simulated computer experiment (Zhang et al., 2001), detecting the impact of autocorrelation on detection and timeliness performance (Lotze and Shmueli, 2009), etc. In this paper, we present a tool of GeneRAting TIme Series with diverse and controllable characteristics, named GRATIS, by exploring the possible time series features and the nature of time dependence. Moreover, in order to show the usefulness of our generation scheme, we develop a novel forecasting selection method based on our generated data in the application section.

Some prior approaches have focused on the shapes of one or more given time series, or on some predefined "types" of time series, in order to generate new time series. Vinod, López-de-Lacalle, et al. (2009) use a maximum entropy bootstrap method to generate ensembles for time series data. The generated samples retain the shape, or local peaks and troughs, of the original time series. They are not exactly the same, but 'strongly dependent', and thus can be used for convenient statistical inference. Bagnall et al. (2017) simulate time series data from different shape settings. The simulators are created by placing one or more shapes on a white noise series. Time series classification algorithms are then evaluated on different representations of the data, which helps understand why one algorithm works better than another on a particular representation of the data. The obvious drawback in these generation methods is that it is impossible to create simulated time series that comprehensively cover the possible space of time series, which limits the reproducibility and applicability of the tested methodology.

An appealing approach is to generate new instances with controllable characteristics, a method that has been used in several other areas of analysis including graph coloring (Smith-Miles and Bowly, 2015), black-box optimization (Muñoz and Smith-Miles, 2017) and machine learning classification (Muñoz et al., 2018). Kang et al. (2017) adapt the idea to time series, and show that it is possible to "fill in" the space of a large collection of real time series data by generating artificial time series with desired characteristics. Each time series is represented using a feature vector which is projected on to a two-dimensional "instance space" that can be visually inspected for diversity. Kang et al. (2017) use a genetic algorithm to evolve new time series to fill in any gaps in the two-dimensional instance space. In a later related paper, Kegel et al. (2017) use STL (an additive decomposition method) to estimate the trend and seasonal component of a series, which they then modify using multiplicative factors to generate new time series. The evolutionary algorithm approach of Kang et al. (2017) is quite general, but computationally slow, while the STL approach of Kegel et al. (2017) is much faster but only generates series that are additive in trend and seasonality. In the meta-learning framework of Talagala et al. (2018), a random forest classifier is used to select the best forecasting method based on time series features. The observed time series are augmented by simulating new time series similar to the observed series, which helps to form a larger dataset to train the model-selection classifier. The simulated series rely on the assumed data generating processes (DGPs), which are exponential smoothing models and ARIMA models.

In this paper, we propose a new efficient and general approach to time series generation, GRATIS, based on Gaussian mixture autoregressive (MAR) models to generate a wide range of non-Gaussian and nonlinear time series. Our generated dataset can be used as benchmarking data in the time series domain, functioning similarly to ImageNet in image processing but with a minimal input of human efforts and computational resources. Mixture transition distribution models were first developed by Le et al. (1996) to capture many general non-Gaussian and nonlinear features, which were later generalized to MAR models (Wong and Li, 2000). We explore generating data from a random population of MAR models, as well as generating data with specified features, which is particularly useful in time series classification or in certain areas where only some features are of interest. In this way, we provide a solution to the need for a heterogeneous set of time series to use for time series analysis.

Finite mixture models have proven useful in many other contexts as well. Different specifications of finite mixtures have been shown to be able to approximate large nonparametric classes of conditional multivariate densities (Jiang and Tanner, 1999; Norets, 2010). More general models to flexibly estimate the density of a continuous response variable conditional on a high-dimensional set of covariates have also been proposed (Li et al., 2010; Villani et al., 2009). Muñoz and Smith-Miles (2017) generate general classification instances with a desired feature vector by fitting Gaussian mixture models. Our GRATIS approach is consistent with this line of literature but we reverse the procedure for generating new time series, in that we use finite mixtures of Gaussian processes to produce time series with specified features.

We first simulate a large time series dataset using MAR models and calculate their features, and the corresponding embedded two-dimensional instance space. The coverage of the simulated data can then be compared with existing benchmarking time series datasets in the two-dimensional space. At the same time, new time series instances with desired features can be generated by tuning a MAR model.

The rest of the paper is organized as follows. Section 2 generates time series from MAR models. Section 3 investigates the diversity and coverage of the generated data in two-dimensional instance spaces. In Section 4, we tune a MAR model efficiently using a genetic algorithm to generate time series data with some specific feature targets. We also study the efficiency of our algorithm. In Section 5, we present a novel time series forecasting approach by exploiting the feature space of generated time series. We show that our scheme can serve as a useful resource for time series applications such as forecasting competitions. Section 6 concludes the paper and discusses possible directions for further studies.

## 2 Time series generation from MAR models

### 2.1 Mixture autoregressive models

Mixture autoregressive models consist of multiple stationary or non-stationary autoregressive components. Being able to capture a great variety of shape-changing distributions, MAR models can handle nonlinearity, non-Gaussianity, cycles and heteroskedasticity in the time series (Wong and Li, 2000). We define a $K$-component MAR model as:

$$F(x_t|\mathcal{F}_{-t}) = \sum_{k=1}^{K} \alpha_k \Phi\left(\frac{x_t - \phi_{k0} - \phi_{k1}x_{t-1} - \cdots - \phi_{kp_k}x_{t-p_k}}{\sigma_k}\right), \qquad (1)$$

where $F(x_t|\mathcal{F}_{-t})$ is the conditional cumulative distribution of $x_t$ given the past information $\mathcal{F}_{-t} \subseteq \{x_{t-1}, \ldots, x_{t-p_k}\}$, $\Phi(\cdot)$ is the cumulative distribution function of the standard normal distribution, $x_t - \phi_{k0} - \phi_{k1}x_{t-1} - \cdots - \phi_{kp_k}x_{t-p_k}$ is the autoregressive term in each mixing component, $\sigma_k > 0$ is the standard error, $\sum_{k=1}^{K} \alpha_k = 1$, and $\alpha_k > 0$ for $k = 1, 2, \ldots, K$. Denoted as MAR($K; p_1, p_2, \ldots, p_k$), it is actually finite mixtures of $K$ Gaussian AR models.

The MAR models have appealing properties as they are based on finite mixture models. For example, the conditional expectation, variance and the $m$th central moment of $x_t$ can be written respectively as:

$$\mathrm{E}(x_t|\mathcal{F}_{-t}) = \sum_{k=1}^{K} \alpha_k(\phi_{k0} + \phi_{k1}x_{t-1} + \cdots + \phi_{kp_k}x_{t-p_k}) = \sum_{k=1}^{K} \alpha_k \mu_{k,t} , \qquad (2)$$

$$\mathrm{Var}(x_t|\mathcal{F}_{-t}) = \sum_{k=1}^{K} \alpha_k \sigma_k^2 + \sum_{k=1}^{K} \alpha_k \mu_{k,t}^2 - \left(\sum_{k=1}^{K} \alpha_k \mu_{k,t}\right)^2, \qquad (3)$$

$$\mathrm{E}((x_t - \mathrm{E}(x_t|\mathcal{F}_{-t}))^m) = \sum_{k=1}^{K} \sum_{i=1}^{m} \alpha_k \binom{m}{i} \mathrm{E}((x_t - \mu_{k,t})^i). \qquad (4)$$

Thus, the MAR model gives a description of the conditional distribution of the time series, and the shape changing feature of the conditional distributions allow the MAR models to describe processes with heteroskedasticity (Wong and Li, 2000). Furthermore, unlike standard AR models, higher order moments (Equation (4)) are also available in MAR densities. To adapt the MAR model to deal with non-stationarity, one can simply include a unit root in each of the $K$ components in Equation (1). Since a seasonal ARIMA model with seasonal effects and unit roots, denoted as ARIMA$(p, d, 0)(P, D, 0)_{\mathsf{Period}}$, can be simply expanded to AR models, one can flexibly include seasonal effects and non-stationarity in any component in the MAR models. Simulating time series with recurrent extreme values is possible when the MAR model has two different components with distinct means (one for normal time and one for extreme time) and unbalanced weights (e.g., 0.99 and 0.01), where the AR process within each component is a seasonal process.

In principle, one can extend the MAR models to mixtures of both autoregressive and moving average models, but we will keep the MAR form as in Equation (1) and not introduce the unnecessary complexity, because both autoregressive and moving average models can be written in terms of autoregressive models. The model in Equation (1) is in univariate form, but it can be extended to the multivariate case by introducing a multivariate normal CDF and vector autoregressive terms. Our approach is fully probabilistic, and incorporates the uncertainty of the parameters in MAR to allow for specific scenarios in time series. This is an alternative to previous studies in Lotze et al. (2010) and Bédubourg and Le Strat (2017) that simulate multivariate time series by adding specific events on simulated baseline data.

In real data, the distribution of the time series can be multi-modal and/or heavy tailed, and so the expectation may not be the best prediction of the future. This is handled nicely with the mixture distribution $F(x_t|\mathcal{F}_{-t})$. From Equation (3), the conditional variance of $x_t$ changes with conditional means of different components. The larger the difference among conditional means $\mu_{k,t}$ ($k = 1, 2, \ldots, K$), the larger the conditional variance of $x_t$. The value of $\sum_{k=1}^{K} \alpha_k \mu_{k,t}^2 - (\sum_{k=1}^{K} \alpha_k \mu_{k,t})^2$ is equal to zero only when $\mu_{1,t} = \mu_{2,t} = \cdots = \mu_{k,t}$ which also yields a heavy-tailed distribution; otherwise, it is larger than zero. The baseline conditional variance is $\sum_{k=1}^{K} \alpha_k \sigma_k^2$.

The key merits of MAR models for nonlinear time series modeling are: (i) for a sufficiently diverse parameters space and finite number of components, MAR models are able to capture extensive time series features in principle (Li et al., 2010), (ii) one can simply include seasonal effects and non-stationary in each component (see Section 2.3), (iii) there is no need to treat stationary and non-stationary time series separately as mixtures of stationary and non-stationary components can yield a both stationary and non-stationary process with MAR (Wong and Li, 2000), (iv) the conditional distributions of the time series given the past information change with time which allows for meaningful time series evolving with historical information, and (v) the MAR models can handle complicated univariate and multivariate time series with different frequencies and seasonalities. The MAR models is also capable of capturing features such as multimodality, heavy tails and heteroskedasticity.

In principle, one may use other types of flexible models as the generator. Nonetheless, our generator based on mixture autoregressive models with minimal parameters setting efficiently generate time series data with diverse features. We describe our time series generation approach and analyze the diversity and coverage of the generated time series in the following sections.

## 2.2 Diverse time series generation

Due to the flexibility of mixture models, they have been successfully applied in many statistical domains such as Bayesian nonparametrics (Escobar and West, 1995), forecasting with high frequency and heavy-tailed time series (Li et al., 2010), model selection (Constantinopoulos et al., 2006) and averaging (Villani et al., 2009), classification methods (Povinelli et al., 2004) and text modeling (Griffiths et al., 2004). Nevertheless, in the extensive literature, little attention has been given to data generation which is crucially important for evaluating the performance of all the tasks mentioned above. Data generating processes do not require sophisticated modeling techniques, but they do require a priori knowledge of the target data space. This space is usually huge and extremely difficult to simulate in a non-time-series context. However, generating diverse time series is possible if one can explore a wide range of time dependencies in time series. In this section we demonstrate how to generate a set of diverse time series data based on the nature of time series dependence.

We design a simulation study to provide insights into the time series simulated from mixture autoregressive models. A significant difference in our data generation process compared to typical simulation processes used in the statistical literature (where the data are generated from models with fixed parameter values), is that we use distributions (see Table 1) instead of fixed values for the parameters in the underlying models. This allows us to generate diverse time series instances. Table 1 shows the parameter settings used in the simulation. These are analogous to non-informative priors (Gelman et al., 2013) in the Bayesian contexts, i.e., the diversity of the generated time series should not rely on the parameter settings.

The periods of the simulated time series are set to be 1, 4, 12 or 52 to match annual, quarterly, monthly and weekly time series. Their lengths are randomly chosen from the lengths of the M4 data (Makridakis et al., 2018). We randomly draw the number of components, $K$, from a uniform distribution on $\{1, 2, 3, 4, 5\}$. Although, it may be desirable to have a larger $K$, Villani et al. (2009) and Li et al. (2010) show that mixture models with comprehensive mean structures are flexible enough with less than five components. The weights of the mixing components, $\alpha_k$, can be obtained as $\beta_k / \sum_{i=1}^{K} \beta_i$ for $k = 1, 2, \ldots, K$, where the $\beta$s follow uniform distributions on (0,1). Assuming that the $k$th component follows a standard seasonal ARIMA model as $\text{ARIMA}(p_k, d_k, 0)(P_k, D_k, 0)_{\text{Period}}$, the coefficients of the AR and seasonal AR parts, $\theta_{ki}$, $i = 1, 2, \ldots, p_k$ and $\Theta_{kj}$, $j = 1, 2, \ldots, P_k$, follow normal distributions with given mean and variance values. In principle, both the coefficients $\theta_{ki}$ and $\Theta_{kj}$ could be unbounded, but this limitation is necessary to keep the features of the simulated data as realistic as possible. For the $k$th mixing component, we perform $d_k$ differences and $D_k$ seasonal differences, where $d_k \sim \text{Bernoulli}(0.9)$ and $D_k \sim \text{Bernoulli}(0.4)$, respectively.

For the parameter settings given in Table 1, we generate 20,000 yearly, 20,000 quarterly, 40,000 monthly and 10,000 weekly time series based on the MAR models. For each generated time series, we discard the first Period $\times$ 10 samples as burn-in. Figure 1 shows examples of simulated yearly, quarterly, monthly and weekly data using GRATIS. The lengths of the simulated series are set to be similar to those of the M4 time series data, which is the largest dataset publicly available to be compared in Section 3. Each of the time series can be summarized with a feature vector described in Section 3.1. For each frequency and each simulated example, we also show (using the same color) the closest real time series from M4 in feature spaces to give an evidence of the realisticity of the generated time series using GRATIS.

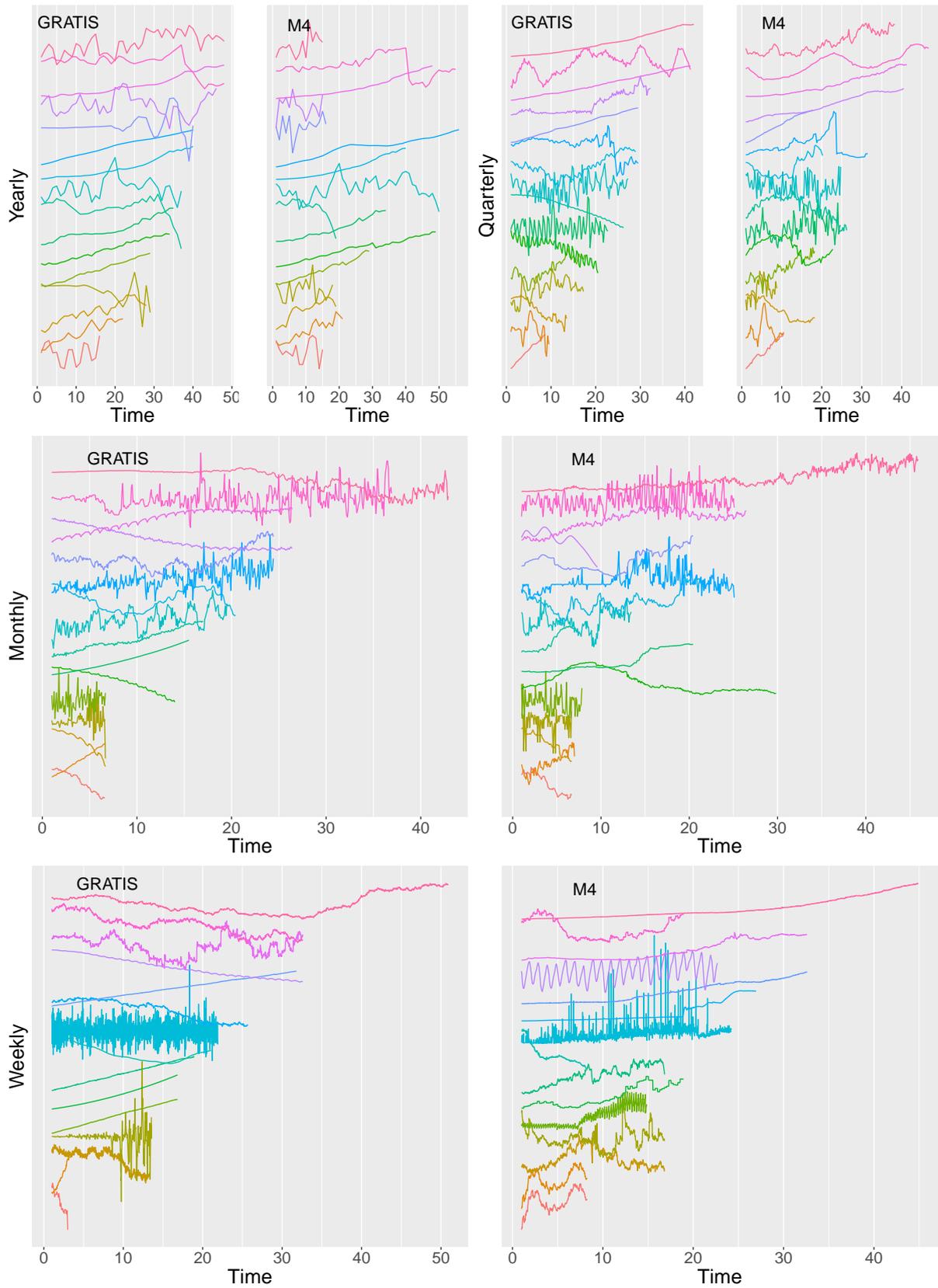

Figure 1: Examples of simulated yearly, quarterly, monthly and weekly time series of different lengths. For each frequency and each simulated example, we also show (using the same color) the closest real time series from M4 in feature spaces. The time series are standardized with centering and scaling and thus the y-axis is omitted. The x-axis for each plot is in years.

Table 1: Parameter settings in our generator for simulating time series using mixtures of ARIMA models.

| Parameter | Description | Values |
|---|---|---|
| Period | Period of time series | 1, 4, 12 or 52 |
| $n$ | Length of time series | Randomly chosen from the lengths of M4 data |
| $K$ | Number of components | $U\{1, 2, 3, 4, 5\}$ |
| $\alpha_k$ | Weights of mixture components | $\alpha_k = \beta_k / \sum_{i=1}^{K} \beta_i$, where $\beta_i \sim U(0, 1)$ |
| $\theta_{ki}$ | Coefficients of the AR parts | $N(0, 0.5)$ |
| $\Theta_{kj}$ | Coefficients of the seasonal AR parts | $N(0, 0.5)$ |
| $\sigma_k$ | Variance of the error terms | $\text{logNormal}(\text{mean} = 0.1, \text{sd} = 0.1)$ |
| $d_k$ | Number of differences in each component | Bernoulli(0.9) |
| $D_k$ | Number of seasonal differences in each component | Bernoulli(0.4) |

Table 2: Computational time (in seconds) for simulation of 1,000 yearly, quarterly, monthly and weekly time series. Different lengths are considered for each seasonal pattern according to the 20%, 50% and 75% quantiles of the time series lengths in M4 data.

| Yearly | | Quarterly | | Monthly | | Weekly | |
|---|---|---|---|---|---|---|---|
| Length | Time (s) | Length | Time (s) | Length | Time (s) | Length | Time (s) |
| 20 | 3 | 60 | 7 | 80 | 13 | 350 | 65 |
| 30 | 3 | 90 | 10 | 200 | 26 | 900 | 156 |
| 40 | 4 | 120 | 13 | 300 | 39 | 1600 | 267 |

Table 2 shows the computational time for simulation of 1,000 yearly, quarterly, monthly and weekly time series. Different lengths are considered for each seasonal pattern according to the 20%, 50% and 75% quantiles of the time series lengths in the M4 data. We have developed an R package `gratis` for the time series generation which is available from https://github.com/ykang/gratis. The code is written in R, and we run it on a laptop with a 2.6 GHz, 8 cores CPU and 16G RAM.

### 2.3 Multi-seasonal time series generation

So far, we have focused on time series in which there is only one seasonal pattern. However, many time series exhibit multiple seasonal patterns of different lengths, especially those series observed at a high frequency (such as daily or hourly data). For example, Figure 2 shows the half-hourly electricity demand for the state of Victoria, Australia, for 5 weeks in late 2014. There is a clear daily pattern of frequency 48, and a weekly pattern of frequency $48 \times 7 = 336$. With a longer time series, an annual pattern would also become obvious.

Simulation of multi-seasonal time series involves weighted aggregation of simulated time series with the corresponding frequencies. A simulated multi-seasonal time series $x_t$ with $M$

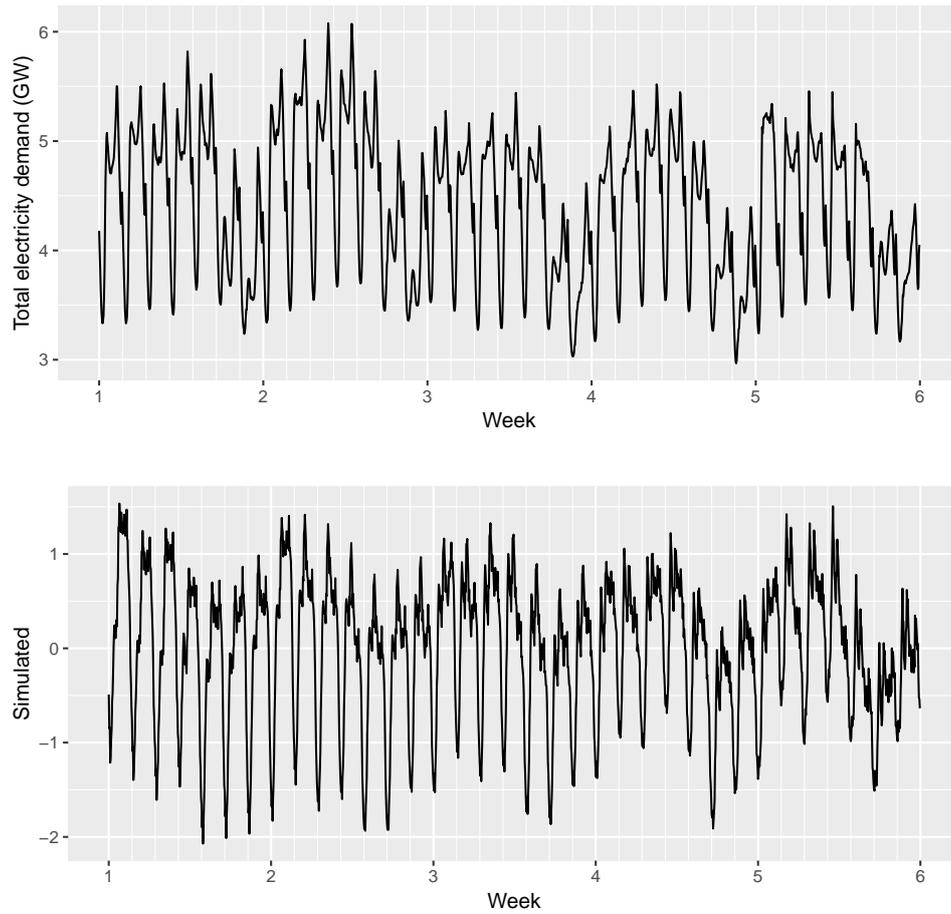

Figure 2: Top panel: Half-hourly electricity demand for Victoria, Australia, with a daily pattern of frequency 48 and a weekly pattern of frequency 336. Bottom panel: Simulated time series with the same seasonal periods and the closest Euclidean distance in features to the real data shown in the top panel.

seasonal patterns can be written as

$$x_t = \sum_{m=1}^{M} \omega_m x_{F_m,t},$$

where $m = 1, 2, \ldots, M$, $x_{F_m,t}$ is the $m$th simulated time series with frequency $F_m$, and weight $\omega_m$ satisfies $\sum_{m=1}^{M} \omega_m = 1$ and $0 < \omega_m < 1$. The weights can be obtained by

$$\omega_m = \frac{\gamma_m}{\sum_{r=1}^{M} \gamma_r}, \text{ where } \gamma_m \sim U(0,1).$$

## 3 Diversity and coverage of generated time series

### 3.1 Time series features

A feature $F_k$ can be any kind of function computed from a time series $\{x_1, \ldots, x_n\}$. Examples include a simple mean, the parameter of a fitted model, or some statistic intended to highlight an attribute of the data.

A unique "best" feature representation of a time series does not exist (Fulcher, 2018). What features are used depends on both the nature of the time series being analyzed, and the purpose of the analysis. For example, consider the mean as a simple time series feature. If some time series contain unit roots, then the mean is not a meaningful feature without some additional constraints on the initial values of the time series. Even if the series are all stationary, if the purpose of our analysis is to identify the best forecasting method, then the mean is probably of no value. On the other hand suppose we are monitoring the CPU usage every minute for numerous servers, and we observe a daily seasonality. Then provided all our time series begin at the same time and are of the same length, the mean provides useful comparative information about the CPU over- or under-utilization in a system despite the time series not being stationary. However, if one is interested in the peak performance of a super computer, the maximum of CPU load is a more desirable feature. This example shows that it is difficult to formulate general desirable properties of features without knowledge of both the time series properties and the required analysis. We encourage analysts using time series features to consider these things before computing many possibly unhelpful or even misleading features.

Because we are studying collections of time series of different lengths, on different scales, and with different properties, we restrict our features to be ergodic, stationary and independent of scale. Specifically, we consider the set of 26 diverse features shown in Table 3. Some features are from previous studies (Wang et al., 2006; Fulcher and Jones, 2014; Kang et al., 2014, 2015; Hyndman et al., 2015; Kang et al., 2017), and some are new features that we believe provide useful information about our data. Our new features are intended to measure attributes associated with multiple seasonality, non-stationarity and heterogeneity of the time series. The features are defined in the Appendix A.1. All features are computed using the `tsfeatures` package (Hyndman et al., 2019) in R (R Core Team, 2018), which nicely handles missing values while calculating the feature vector for a given time series.

Table 3: The features we use to characterize a time series.

| Feature | Name | Description | Range |
| --- | --- | --- | --- |
| $F_1$ | length | Length of the time series | $[1, \infty)$ |
| $F_2$ | nPeriods | Number of seasonal periods | $[1, \infty)$ |
| $F_3$ | Periods | Vector of seasonal periods | $\{1, 2, 3, \dots\}$ |
| $F_4$ | ndiffs | Number of differences for stationarity | $\{0, 1, 2, \dots\}$ |
| $F_5$ | nsdiffs | Number of seasonal differences for stationarity | $\{0, 1, 2, \dots\}$ |
| $F_6$ | (x.acf1, x.acf10, diff1.acf1, diff1.acf10, diff2.acf1, diff2.acf10, seas.acf1) | Vector of autocorrelation coefficients | (-1, 1) or $(0, \infty)$ |
| $F_7$ | (x.pacf5, diff1.pacf5, diff2.pacf5, seas.pacf) | Vector of partial autocorrelation coefficients | (-1, 1) or $(0, \infty)$ |
| $F_8$ | entropy | Spectral entropy | (0, 1) |
| $F_9$ | nonlinearity | Nonlinearity coefficient | $[0, \infty)$ |
| $F_{10}$ | hurst | Long-memory coefficient | [0.5, 1] |
| $F_{11}$ | stability | Stability | $(0, \infty)$ |
| $F_{12}$ | lumpiness | Lumpiness | $[0, \infty)$ |
| $F_{13}$ | (unitroot.kpss, unitroot.pp) | Vector of unit root test statistics to indicate the strength of non-stationarity | $(0, \infty)$ or $(-\infty, \infty)$ |
| $F_{14}$ | (max.level.shift, time.level.shift) | Maximum level shift | $(0, \infty)$ |
| $F_{15}$ | (max.var.shift, time.var.shift) | Maximum variance shift | $(0, \infty)$ |
| $F_{16}$ | (max.kl.shift, time.kl.shift) | Maximum shift in Kulback-Leibler divergence | $(0, \infty)$ |
| $F_{17}$ | trend | Strength of trend | [0, 1) |
| $F_{18}$ | seasonal.strength | Strength of seasonality | [0, 1) |
| $F_{19}$ | spike | Spikiness | [0, 1) |
| $F_{20}$ | linearity | Linearity | $(-\infty, \infty)$ |
| $F_{21}$ | curvature | Curvature | $(-\infty, \infty)$ |
| $F_{22}$ | (e.acf1, e.acf10) | Vector of autocorrelation coefficients of remainder | (-1, 1) or $(0, \infty)$ |
| $F_{23}$ | arch.acf | Heterogeneity measure by ARCH ACF statistic to indicate ARCH effects | $(0, \infty)$ |
| $F_{24}$ | garch.acf | Heterogeneity measure by GARCH ACF statistic to indicate GARCH effects | $(0, \infty)$ |
| $F_{25}$ | arch.r2 | Heterogeneity measure by ARCH $R^2$ statistic to indicate ARCH effects | [0, 1] |
| $F_{26}$ | garch.r2 | Heterogeneity measure by GARCH $R^2$ statistic to indicate GARCH effects | [0, 1] |

Little previous study has used features for multiple seasonal time series. In multiple seasonal time series, there is more than one seasonal period present in the data; for example, hourly electricity demand data contains a time-of-day pattern (with seasonal period 24), a time-of-week pattern (with seasonal period $7 \times 24 = 168$) and a time-of-year pattern (with seasonal period $365 \times 24 = 8760$). If there are $M$ possible seasonal periods, then $F_2 = M$ and $F_3$ is an $M$-vector containing the seasonal periods. For example, with monthly data $F_3 = 12$, and with hourly data $F_3 = (24, 168, 8760)'$. The strength of seasonality ($F_{18}$) is also an $M$-vector

Table 4: Benchmarking datasets used for comparison with the simulated series from MAR models. The number of series is shown per dataset and seasonal pattern.

| Dataset | R package | Yearly | Quarterly | Monthly | Weekly | Daily |
|---------|-----------|--------|-----------|---------|--------|-------|
| M1      | Mcomp     | 181    | 203       | 617     | –      | –     |
| M3      | Mcomp     | 645    | 756       | 1428    |        |       |
| M4      | M4comp2018| 23000  | 24000     | 48000   | 359    | 4227  |
| Tourism | Tcomp     | 518    | 427       | 366     | –      | –     |
| NNGC1   | tscompdata| 11     | 11        | 11      | 11     | 11    |

containing separate measures of the strength of seasonality for each of the seasonal periods.

We investigate the diversity and coverage of the generated time series data based on MAR models by comparing the feature space of the simulated data with feature spaces of several benchmarking time series datasets, including those from the M1, M3, M4, Tourism and NNGC1 forecasting competitions. The R packages they come from, and the numbers of yearly, quarterly, monthly, weekly and daily time series in each dataset are shown in Table 4.

## 3.2 Diversity and coverage analysis

First, we analyze the feature diversity from a marginal perspective. Figure 3 depicts the feature diversity and coverage for simulated yearly, quarterly, monthly and weekly time series compared to the benchmarks for all the possible features we use in the paper. The features in our generated data are diverse in the sense that (i) the shapes of the feature plots for the simulated data widely match to the theoretical ranges of features given in Table 3, and (ii) the quantiles of the features for the simulated data cover the shapes of all features in the benchmarking data.

To better understand the feature space, Kang et al. (2017) use principal component analysis (PCA) to project the features to a 2-dimensional "instance space" for visualization. A major limitation of any linear dimension reduction method is that it puts more emphasis on keeping dissimilar data points far apart in the low dimensional space. But in order to represent high dimensional data in a low dimensional, nonlinear manifold, it is also important that similar data points are placed close together. When there are many features and nonlinear correlations are present, using a linear transformation of the features may be misleading. Therefore, we use t-Stochastic Neighbor Embedding (t-SNE), a nonlinear technique capable of retaining both local and global structure of the data in a single map, to conduct the nonlinear dimension reduction of the high dimensional feature space (Maaten and Hinton, 2008).

Figure 4 shows a comparison of PCA and t-SNE for the M3 data. The top row of the plot shows the distribution of the seasonal period `period` in the t-SNE and PCA spaces, while the bottom row shows that of the spectral entropy feature `entropy`. It can be seen that t-SNE is better able to capture the nonlinear structure in the data, while the linear transformation of PCA leads to a large proportion of information loss, especially when more features are used. The distribution of other features can be studied similarly.

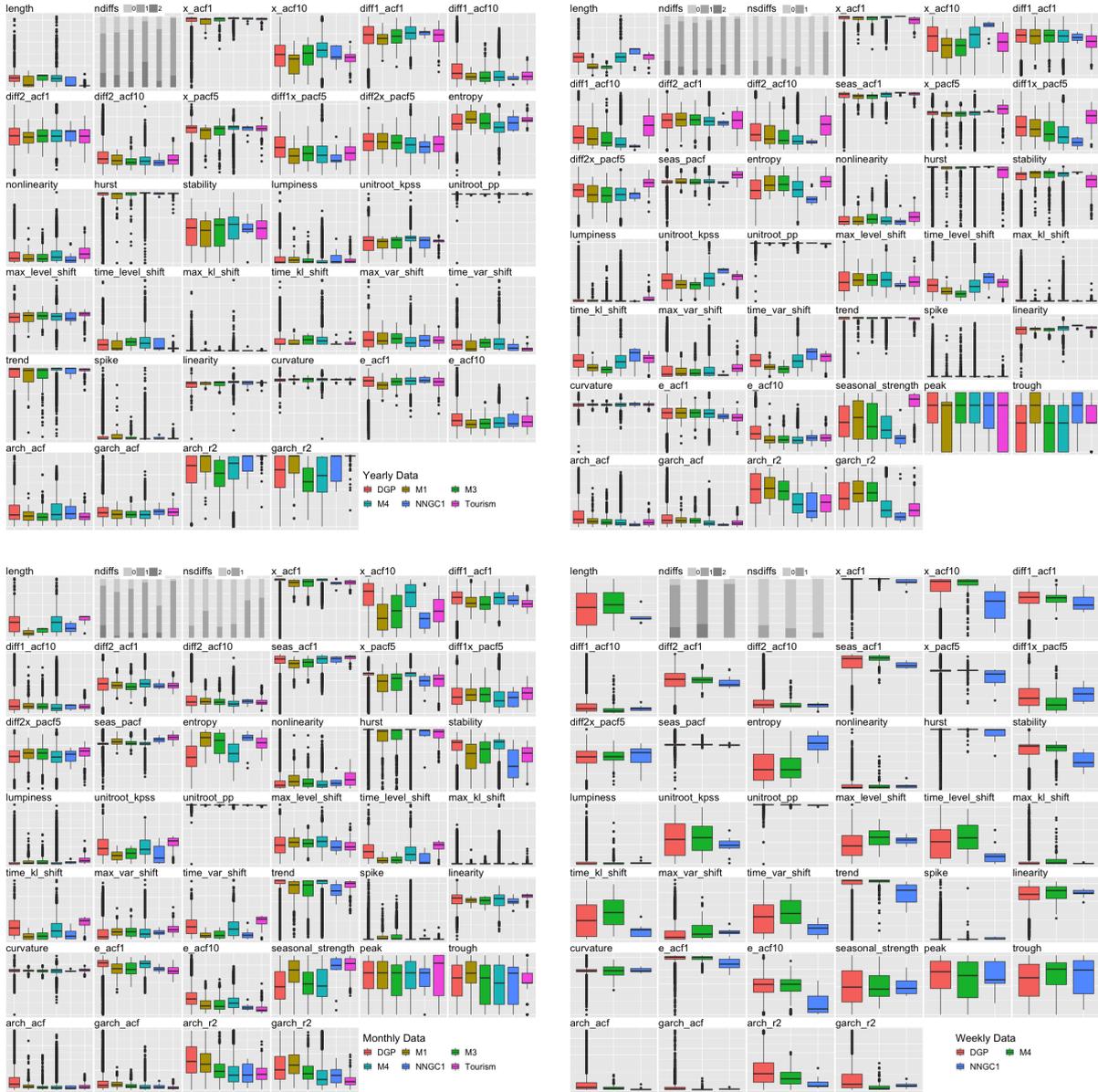

Figure 3: Plots showing feature diversity and coverage of the simulated yearly, quarterly, monthly, and weekly time series compared with the benchmarking data M1, M3, M4, NNGC1 and Tourism. Boxplots are used for features with continuous values while percentage bar charts for discrete cases. In all the plots, the same order of the datasets is used.

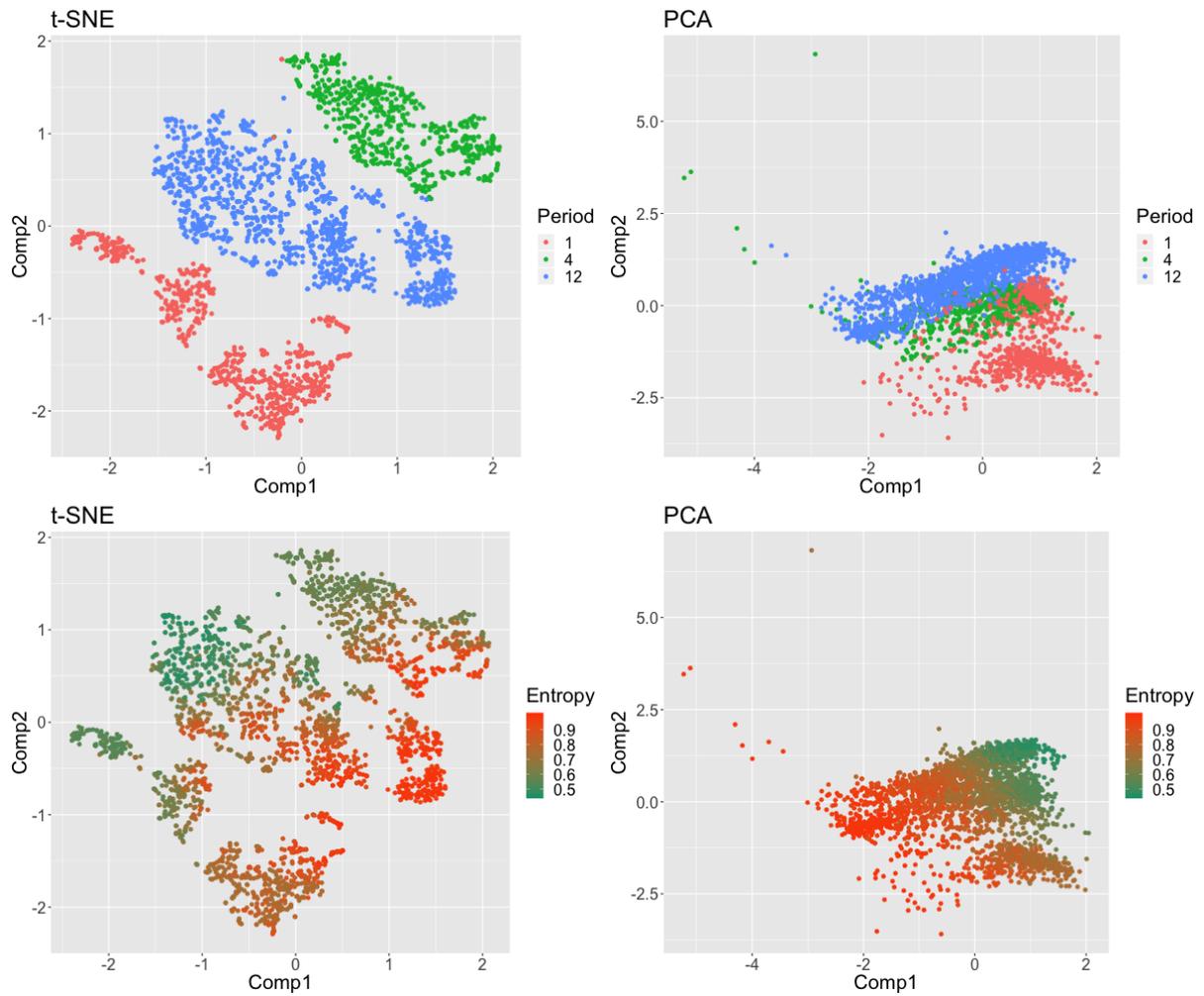

Figure 4: Two-dimensional spaces of M3 data based on t-SNE (left) and PCA (right). Comp1 and Comp2 are the first two components after dimension reduction using t-SNE and PCA. The top row shows the distribution of period in the two spaces, while the bottom row shows that of entropy.

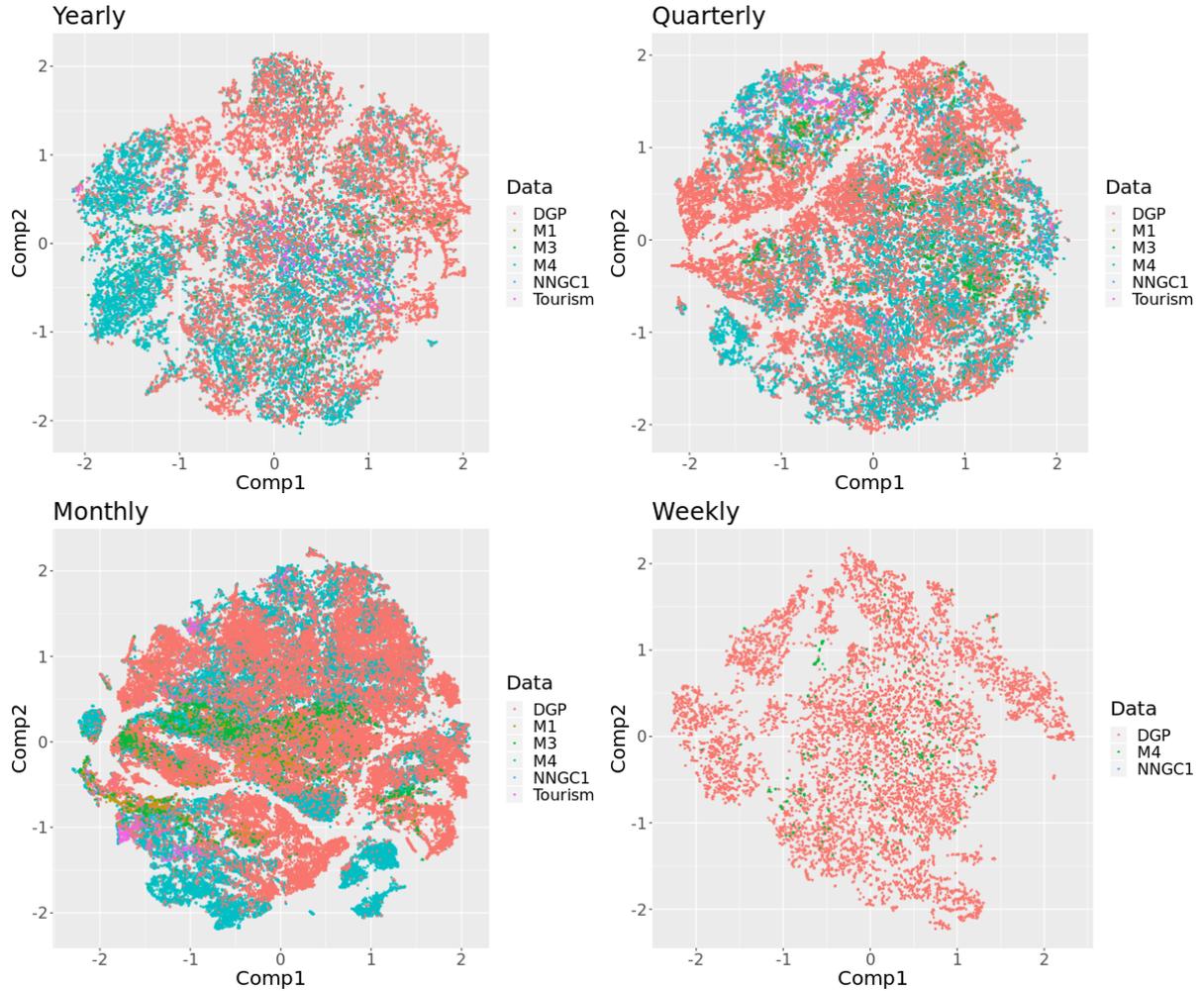

Figure 5: Two-dimensional t-SNE spaces of the simulated yearly, quarterly, monthly, and weekly time series, together with the time series with the same seasonal patterns from the M1, M3, M4, NNGC1 and Tourism datasets. Comp1 and Comp2 are the first two components after dimension reduction using t-SNE.

The simulated yearly, quarterly, monthly and weekly time series are projected into a two-dimensional feature space together with the yearly, quarterly, monthly and weekly time series in the benchmarking datasets, as shown in Figure 5. Time series with different seasonal patterns are shown in separate panels of Figure 5 to make the comparisons easier.

Given the two-dimensional feature spaces of dataset A and dataset B, we quantify the miscoverage of dataset A over dataset B in the following steps:

1. Find the maximum ranges of the $x$ and $y$ axes reached by the combined datasets $A$ and $B$, and cut the $x$ and $y$ dimensions into $N_b = 30$ bins.

2. In the constructed two-dimensional grid with $N_b^2 = 900$ subgrids, we denote

$$\mathcal{I}_{i,A} = \begin{cases} 0 & \text{if no time series in dataset A fall into the } i\text{th subgrid;} \\ 1 & \text{otherwise.} \end{cases}$$

An analogous definition of $\mathcal{I}_{i,B}$ applies for dataset B.

3. The miscoverage of dataset $A$ over dataset $B$ is defined as

$$\text{miscoverage}_{A/B} = N_b^{-2} \sum_{i=1}^{N_b} [(1 - \mathcal{I}_{i,A}) \times \mathcal{I}_{i,B}]. \tag{5}$$

One could use $(1 - \text{miscoverage}_{A/B})$ to infer the relative diversity of dataset $A$ over the dataset $B$. Specifically, if dataset $A$ is the simulated time series and dataset $B$ is a rich collection of real time series, $(1 - \text{miscoverage}_{A/B})$ can also be used as a measure of how realistic of the simulated time series set $A$ by taking the real dataset $B$ as the ground truth.

Table 5 shows the pairwise miscoverage values of benchmarking dataset A over B. Again, time series with different seasonal patterns are shown separately. The miscoverage values of the simulated dataset from the DGP over others are always smaller than that of others over the DGP. Focusing on the M4 data, the most comprehensive time series competition data to date, the miscoverage values of the DGP over M4 are 0.017, 0.013, 0.030 and 0.001 for yearly, quarterly, monthly and weekly data, respectively. On the other hand, the miscoverage values of M4 over DGP are substantially higher. Therefore, together with Figure 5, we find that the simulated data from MAR models bring more diversity than the existing benchmarking datasets. Therefore, Figure 5 and Table 5 present two useful tools for a researcher to screen the diversity of simulated time series and their realisticity compared to a collection of real time series.

The coverage analysis indicates the possibility of generating time series in a certain feature space. From Figure 5, one may notice that some areas of the feature space contain more generated time series than other areas with the default parameter settings in Table 1. Possible ways to allow the time series to be more evenly distributed in the feature space are either to adapt the parameter setting in Table 1 to make the generator focus more on the sparse area, or enlarge the number of simulations and incorporate with an accept-reject scheme, or use the time series generation method with target features proposed in Section 4.

Table 5: Miscoverage of dataset A over dataset B. Take the yearly section for example, the miscoverage of simulated data over M4 is 0.017, while the miscoverage of M4 over simulated data is 0.053.

| Dataset A | Dataset B | | | | | |
|---|---|---|---|---|---|---|
| | DGP | M4 | M3 | M1 | Tourism | NNGC1 |
| | | | Yearly | | | |
| DGP | 0.000 | 0.017 | 0.001 | 0.001 | 0.000 | 0.001 |
| M4 | 0.053 | 0.000 | 0.001 | 0.001 | 0.000 | 0.000 |
| M3 | 0.609 | 0.550 | 0.000 | 0.041 | 0.106 | 0.008 |
| M1 | 0.680 | 0.629 | 0.113 | 0.000 | 0.107 | 0.008 |
| Tourism | 0.622 | 0.568 | 0.108 | 0.033 | 0.000 | 0.007 |
| NNGC1 | 0.720 | 0.669 | 0.126 | 0.052 | 0.124 | 0.000 |
| | | | Quarterly | | | |
| DGP | 0.000 | 0.013 | 0.000 | 0.000 | 0.000 | 0.000 |
| M4 | 0.079 | 0.000 | 0.001 | 0.000 | 0.001 | 0.000 |
| M3 | 0.567 | 0.536 | 0.000 | 0.043 | 0.096 | 0.009 |
| M1 | 0.651 | 0.616 | 0.149 | 0.000 | 0.110 | 0.009 |
| Tourism | 0.660 | 0.621 | 0.200 | 0.112 | 0.000 | 0.010 |
| NNGC1 | 0.769 | 0.729 | 0.243 | 0.132 | 0.137 | 0.000 |
| | | | Monthly | | | |
| DGP | 0.000 | 0.030 | 0.001 | 0.000 | 0.002 | 0.000 |
| M4 | 0.061 | 0.000 | 0.003 | 0.001 | 0.000 | 0.000 |
| M3 | 0.488 | 0.446 | 0.000 | 0.020 | 0.052 | 0.002 |
| M1 | 0.566 | 0.523 | 0.131 | 0.000 | 0.063 | 0.004 |
| Tourism | 0.654 | 0.602 | 0.251 | 0.162 | 0.000 | 0.006 |
| NNGC1 | 0.716 | 0.669 | 0.288 | 0.204 | 0.084 | 0.000 |
| | | | Weekly | | | |
| DGP | 0.000 | 0.001 | – | – | – | 0.000 |
| M4 | 0.456 | 0.000 | – | – | – | 0.007 |
| M3 | – | – | – | – | – | – |
| M1 | – | – | – | – | – | – |
| Tourism | – | – | – | – | – | – |
| NNGC1 | 0.576 | 0.138 | – | – | – | 0.000 |

# 4 Efficient time series generation with target features

## 4.1 Tuning a MAR model with target features

In time series analysis, researchers with a particular focus may be only interested in a certain area of the feature space, or a subset of features, e.g., heteroscedasticity and volatility in financial time series, trend and entropy in time series forecasting, or peaks and spikes in energy time series. Practitioners may also want to mimic more time series from their limited collection of real data such as sales records for new products and health metrics for a rare disease. Therefore, efficient generation of time series with target features of interest is another important problem to address. For a review of the relevant literature, see Kang et al. (2017).

Kang et al. (2017) use a genetic algorithm (GA) to evolve time series of length $n$ that project to the target feature point $\tilde{\boldsymbol{F}}$ in the two-dimensional instance space as closely as possible. At each iteration of the GA, a combination of selection, crossover and mutation is applied over the corresponding population to optimize the $n$ points of the time series to be evolved. The computational complexity grows linearly as the length of time series increases. We document the details of the genetic algorithm in Appendix A.4.

In this paper, instead of evolving time series of length $n$ (i.e., optimization in an $n$-dimension space), we use a GA to tune the MAR model parameters until the distance between the target feature vector and the feature vector of a sample of time series simulated from the MAR is close to zero. The parameters for the MAR model, the underlying DGP, can be represented as a vector $\Theta = \{\alpha_k, \phi_i\}$ for $k = 1, \ldots, K$ and $i = k0, \ldots, kp_k$ in Equation (1). A significant improvement compared to Kang et al. (2017) is that the length of the vector $\Theta$ is much smaller than $n$, so that the tuning process can be performed efficiently in a much lower-dimensional parameter space. The GA optimization steps are summarized below, for a specified period Period and length $n$ for the desired time series.

1. Select a target feature point $\tilde{\boldsymbol{F}}$ in the feature space or let $\tilde{\boldsymbol{F}}$ be the extracted features from existing time series that one wants to match. Now we aim to find a parameter vector $\Theta^*$ that can evolve a time series $X_{\tilde{\boldsymbol{F}}}$ with its feature vector $\boldsymbol{F}$ as close as possible to the target feature point $\tilde{\boldsymbol{F}}$.

2. Generate an initial population of size $N_P$ for the parameter vector $\Theta$, in which each parameter is chosen from a uniform distribution as given in Table 1. That allows the entire range of possible solutions of $\Theta$ to be reached randomly.

3. For each iteration, repeat the following steps until some stopping criteria are met.

(a) For each member in the current population, simulate a time series $j$ and calculate its feature vector $\boldsymbol{F}_j$.

(b) Calculate the fitness value for each member:

$$\text{Fitness}(j) = -\frac{1}{c}\|\boldsymbol{F}_j - \tilde{\boldsymbol{F}}\|,$$

where $c$ is a scaling constant usually defined as $c = \|\tilde{\boldsymbol{F}}\|$ and $\|\cdot\|$ is a distance measure for the feature space. We find that Euclidean distance works well in our algorithm.

(c) Produce the new generation based on the crossover, mutation and the survival of the fittest individual to improve the average fitness value of each generation.

4. Keep the time series that is closest to the target feature point (i.e., has the largest fitness value) to be the newly generated time series for the corresponding target.

One may want to mimic very short time series. In principle, GRATIS could be equally well used if the features of target series are available. Nonetheless, if the feature calculation involves lagged values, the length of time series should be at least greater than the length of lags. Besides, expert information can be also provided through the settings in Table 1, which is particularly useful when one has very limited historical data or one is interested in a particular setting of features.

## 4.2 Efficiency analysis

Table 6 shows the computational time for generating different time series with different sets of features. Lengths are chosen in the same way as for Table 2. The target feature vectors used are median values of the selected features of the simulated data with the same seasonal pattern and similar lengths. The algorithm used in Kang et al. (2017) takes about 22,000 seconds to evolve 100 time series of length 100 with 6 features. For this similar task, but with twice as many features, our algorithm is about 40 times faster on average. This speedup allows us to generate time series with controllable features in a reasonable time.

Table 6: Computational time (in seconds) for generation of 100 yearly, quarterly, monthly and weekly time series. Feature set A consists of ndiffs, x_acf1, entropy and trend. Feature set B consists of Feature set A, diff1_acf1, seasonal_strength, seas_pacf and e_acf1. Feature set C consists of Feature set B, e_acf10, unitroot_kpss, linearity and garch_r2. Median values of the selected features of the simulated data with the same seasonal pattern and similar lengths are used as the targets.

| Yearly | | Quarterly | | Monthly | | Weekly | |
|---|---|---|---|---|---|---|---|
| Length | Time (s) | Length | Time (s) | Length | Time (s) | Length | Time (s) |
| Feature set A (four features) | | | | | | | |
| 20 | 53 | 60 | 78 | 80 | 62 | 350 | 124 |
| 30 | 40 | 90 | 71 | 200 | 74 | 900 | 182 |
| 40 | 40 | 120 | 87 | 300 | 132 | 1600 | 265 |
| Feature set B (eight features) | | | | | | | |
| 20 | 43 | 60 | 130 | 80 | 524 | 350 | 3001 |
| 30 | 119 | 90 | 319 | 200 | 480 | 900 | 3395 |
| 40 | 101 | 120 | 405 | 300 | 1340 | 1600 | 3674 |
| Feature set C (twelve features) | | | | | | | |
| 20 | 180 | 60 | 650 | 80 | 1190 | 350 | 1655 |
| 30 | 550 | 90 | 530 | 200 | 349 | 900 | 1360 |
| 40 | 202 | 120 | 1160 | 300 | 725 | 1600 | 1573 |

## 5 Application to time series forecasting

In general, our time series generation scheme, GRATIS, can serve as a useful resource for various advanced time series analysis, such as time series forecasting and classification. For illustration purposes, we present a novel time series forecasting approach by exploiting the generated time series. It is worth mentioning that the construction of such a large-scale and high-quality database is efficient and does not rely on traditional data collection methods.

### 5.1 Forecasting based on feature spaces in generated time series

The No-Free-Lunch theorem states there is never universally a best method that fits in all situations in machine learning and optimization (Wolpert, 1996; Wolpert and Macready, 1997). This idea also applies to the context of time series forecasting, in which no single forecasting method stands out as best for any type of time series, e.g., (Adam, 1973; Collopy and Armstrong, 1992; Wang et al., 2009; Petropoulos et al., 2014; Kang et al., 2017). Adam (1973) showed that the statistical characteristics of each time series are related to the accuracy of each forecasting method. Later literature, e.g., (Adam, 1973; Shah, 1997; Meade, 2000) tend to support this argument. An ideal case is that one can select the best forecast model for each series in a dataset according to its features. Another way is to calculate the averaging weights of the forecasts from individual forecast models for each time series based on its features.

We aim to examine how those features influence forecasting method performance through simulations from mixture autoregressive models, which enables us to predict the performances of the forecasting methods on the candidate time series data, and select the best forecasting method or average different forecasting methods with proper weights. We describe the complete diagram in Figure 6. The task details are partitioned into training and testing procedures as below.

1. Training on generated time series:

   (a) Simulate 10,000 yearly, quarterly and monthly time series as the training time series with GRATIS. The forecasting horizon $h$ are set as 6, 8, 18 for yearly, quarterly and monthly data, respectively.

   (b) Calculate the training feature matrix $F^{(train)}$ for the historical data.

   (c) Calculate the out-of-sample $MASE^{(train)}$ (Mean Absolute Scaled Error; Hyndman and Koehler, 2006) values on the forecasting horizon for nine commonly used time series forecasting methods.

   (d) Model the relationship between the feature matrix $F^{(train)}$ and $MASE^{(train)}$ with nonlinear regression models.

2. Forecasting on testing time series:

   (a) Calculate the feature matrix $F^{(test)}$.

   (b) Predict the best forecasting method that minimizes the predicted $\widehat{MASE}^{(test)}$ for each time series. Or use a model averaging procedure, that is, calculate the weight of $m$th model ($\vartheta_m$) as

   $$\vartheta_m = \frac{\exp{(1/\widehat{MASE}_k^3)}}{\sum_{m=1}^{M} \exp{(1/\widehat{MASE}_m^3)}}.$$

   where $\widehat{MASE}_m$ is the predicted MASE value from the nonlinear regression model in Figure 6.

We use MASE as the measure of forecasting accuracy because it is stated in Hyndman and Koehler (2006) as a " generally applicable measurement of forecast accuracy without the problems seen in the other measurements". It has also been shown that MASE does not substantially affect the main conclusions about the best-performing methods in the M3 competition data (Hyndman and Koehler, 2006). It is worth mentioning that, in principle, our method is general and can be incorporated with any other forecasting measures. All steps except step 1.(d) can be parallelized to reduce the computation burden and our R package `gratis` is designed for parallel computing. In step 1.(d), we find that quantile regression with a lasso penalty works the best for seasonal data, i.e., quarterly and monthly data. We document the model details in Appendix A.3, whilst multivariate adaptive spline regression models (Friedman et al., 1991) work well for non-seasonal data, i.e., yearly data in our case.

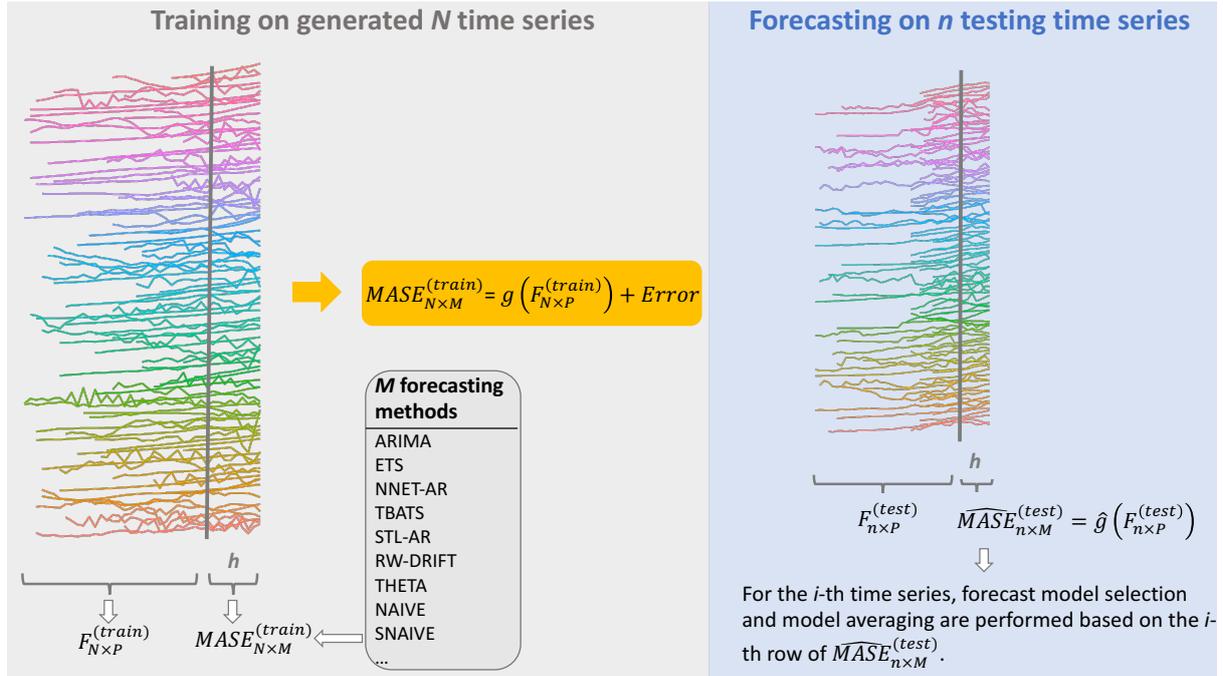

Figure 6: Forecasting diagram based on generated time series. The notion $N$ is the number of training time series, $P$ is the number of time series features, $M$ is the number of forecasting methods, $n$ is the number of testing series, and $g(\cdot)$ describes the nonlinear relationship between $MASE$ values and time series features.

This modeling approach with GRATIS is novel in several ways: (i) the training process is done purely on generated time series and does not involve the candidate time series, (ii) it does not require us to try all potential models on the testing time series which makes it highly computationally efficient when the testing data is very large, (iii) it does not require real data input in the training process, and our trained model can serve as a general pre-trained time series forecasting algorithm selector, and (iv) most importantly, the whole procedure only requires the transformed nonsensitive time series features instead of real time series as the forecasting input. This is particularly useful when privacy is a concern.

## 5.2 Application to M3 data

For demonstration purposes, we apply our pretrained models to the yearly, quarterly and monthly data in M3. We consider nine commonly used forecasting methods (Hyndman and Athanasopoulos, 2018): automated ARIMA algorithm (ARIMA), automated exponential smooth-ing algorithm (ETS), NNET-AR model applying a feed-forward neural network using autoregressive inputs (NNET-AR), TBATS model (Exponential Smoothing State Space Model With Box-Cox Transformation, ARMA Errors, Trend And Seasonal Components), Seasonal and Trend decomposition using Loess with AR modeling of the seasonally adjusted series (STL-AR), random walk with drift (RW-DRIFT), theta method (THETA), naive (NAIVE), and seasonal naive (SNAIVE).

Table 7: Comparison of the MASE values from feature-based forecasting using GRATIS and the other individual nine methods on the M3 data. Simple averaging shows the MASE values based on the mean of the nine forecast values from the individual methods.

| Forecasting Method | Yearly | Quarterly | Monthly | Overall |
|---|---|---|---|---|
| | | Benchmarking | | |
| ARIMA | 1.898 | 0.841 | 0.710 | 0.855 |
| ETS | 1.907 | 0.855 | 0.712 | 0.869 |
| NNET-AR | 2.338 | 1.021 | 0.828 | 1.037 |
| TBATS | 1.900 | 0.914 | 0.699 | 0.869 |
| STLM-AR | 2.625 | 1.688 | 1.010 | 1.326 |
| RW-DRIFT | 1.929 | 0.988 | 0.894 | 1.046 |
| THETA | 1.985 | 0.831 | 0.721 | 0.869 |
| NAIVE | 2.267 | 1.044 | 0.927 | 1.135 |
| SNAIVE | 2.267 | 1.176 | 0.969 | 1.146 |
| Simple Averaging | 1.964 | 0.834 | 0.736 | 0.872 |
| | | GRATIS | | |
| Model Selection | 1.713 | 0.792 | 0.694 | 0.831 |
| Model Averaging | 1.759 | 0.784 | 0.688 | 0.807 |

For each time series, forecasting method selection is performed according to the smallest predicted MASE values. For comparison purposes, we also calculate the MASE values for all nine forecasting algorithms applied to each of M3 time series. Table 7 depicts the median of the MASE values for our method selection and averaging procedures, and the nine individual methods on each group of time series in M3. The MASE values from model selection show that our feature-based forecasting scheme trained with generated time series gives the lowest forecasting MASE in all individual groups and the overall dataset. Under the model averaging scheme, our feature-based forecasting scheme trained with generated time series further increases forecasting accuracy in terms of MASE. Table 8 shows the model selection proportions and model averaging weights on M3 with GRATIS.

Figure 7 depicts the distribution of MASE with boxplots for yearly, quarterly, monthly and overall data in M3. Our proposed method is not only the best on average but also avoids very large deviations, compared to other nine methods. Figure 8 shows the projection of the best predicted forecasting method by our feature-based model selection algorithm on two-dimensional feature spaces for the yearly, quarterly, monthly data in M3, from which one may observe distinct patterns exist. For example, random walk with drift (RW-DRIFT) performs the best for the quarterly time series lying in the middle of the feature space, while ARIMA and TBATS models works well for the monthly data lying in the left region of the feature space.

Table 8: Model selection proportions and model averaging weights on M3 with GRATIS. In each column, "%" means the percentage of times the corresponding method is selected in feature-based model selection and $\vartheta$ means the median weight for each method used in feature-based model averaging.

|  | Yearly | | Quarterly | | Monthly | | Overall | |
| --- | --- | --- | --- | --- | --- | --- | --- | --- |
| Forecasting Method | Selection (%) | Averaging ($\vartheta$) | Selection (%) | Averaging ($\vartheta$) | Selection (%) | Averaging ($\vartheta$) | Selection (%) | Averaging ($\vartheta$) |
| ARIMA | 26.8 | 0.157 | 41.3 | 0.180 | 55.6 | 0.188 | 45.2 | 0.180 |
| ETS | 14.3 | 0.131 | 7.1 | 0.155 | 4.3 | 0.138 | 7.4 | 0.139 |
| NNET-AR | 1.2 | 0.064 | 1.3 | 0.060 | 0.7 | 0.096 | 1.0 | 0.076 |
| TBATS | 9.9 | 0.128 | 5.3 | 0.139 | 9.0 | 0.130 | 8.2 | 0.131 |
| STLM-AR | 3.1 | 0.045 | 1.2 | 0.024 | 4.6 | 0.058 | 3.3 | 0.045 |
| RW-DRIFT | 20.5 | 0.131 | 29.1 | 0.138 | 13.9 | 0.077 | 19.4 | 0.106 |
| THETA | 16.7 | 0.120 | 5.4 | 0.087 | 2.0 | 0.084 | 6.3 | 0.092 |
| NAIVE | 7.4 | 0.078 | 7.9 | 0.080 | 6.6 | 0.080 | 7.1 | 0.080 |
| SNAIVE | – | 0.078 | 1.3 | 0.057 | 3.4 | 0.064 | 2.1 | 0.065 |

It is worth mentioning that although our generated time series dataset is based on mixtures of autoregressive models, the ARIMA models are not always selected as the best in the generated data, providing further evidence that our generated time series are diverse. This is due to the flexibility of mixtures and our random parameter settings in the generator in Table 1. In order to gain further insight into our forecasting approach, we fit a MAR to each candidate series of the yearly data from M3 and estimate the model parameters using maximum likelihood estimation. The median of the resulting MASE is 2.311 which is much greater than the MASE values from model selection and model averaging in Table 7. This indicates that MAR models may not be adequate to model each series, but they are capable of producing a suitably rich collection of time series to allow one to predict when a given forecasting method is likely to perform well according to the time series characteristics.

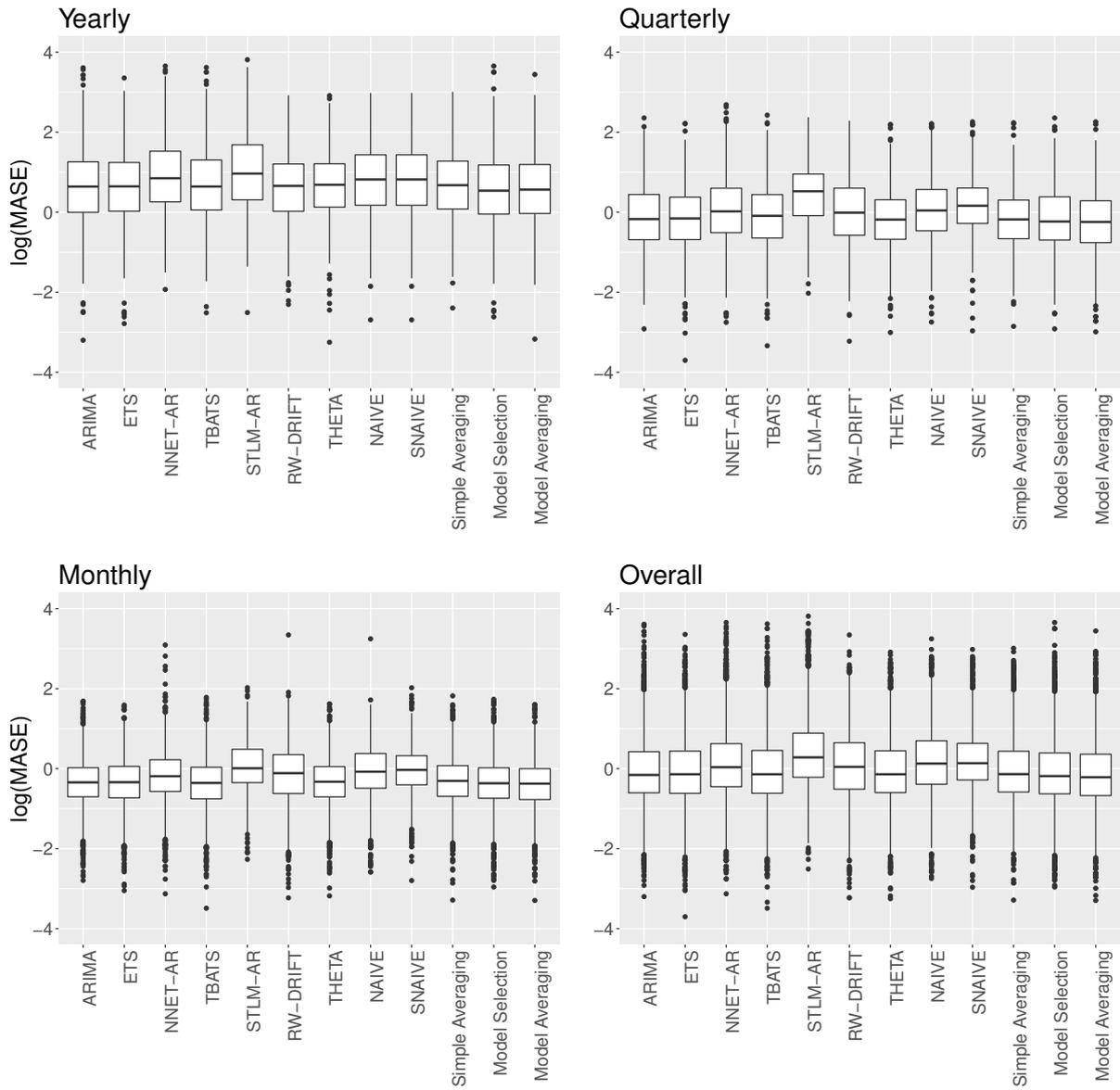

Figure 7: Boxplot of log scaled MASE values for yearly, quarterly, monthly and overall data in M3.

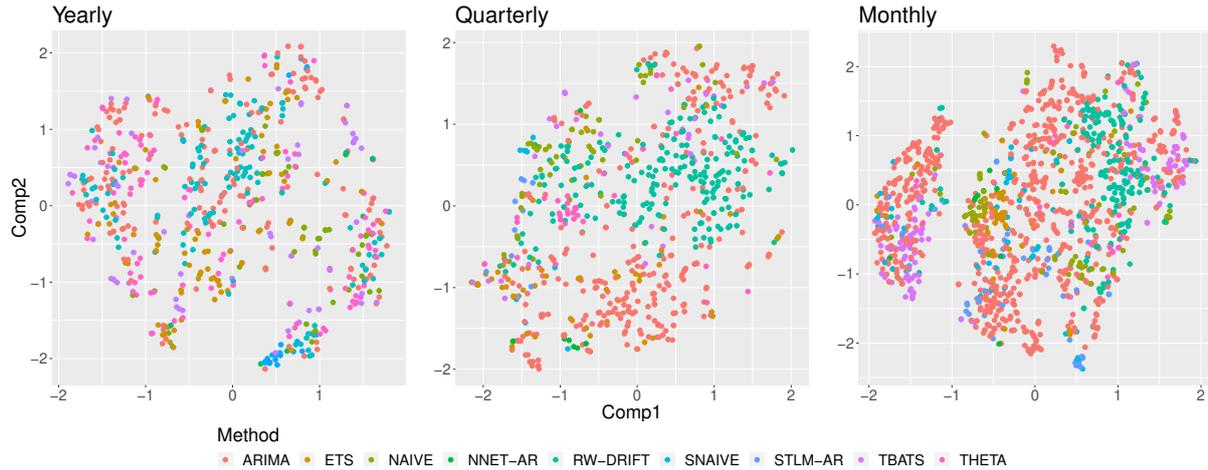

Figure 8: Projection of best predicted forecasting method by our feature-based model selection algorithm on two-dimensional feature spaces for the yearly, quarterly, monthly data in M3.

# 6 Conclusions and discussion

We have proposed an efficient simulation method, GRATIS, for generating time series with diverse characteristics requiring minimal input of human effort and computational resources. Our generated dataset can be used as benchmarking data in the time series domain, which functions similarly to other machine learning data repositories. The simulation method is based on mixture autoregressive models where the parameters are assigned with statistical distributions. In such a way, we provide a general benchmarking tool serving for advanced time series analysis where a large collection of benchmarking data is required, including forecasting comparison, model averaging, and time series model training with self-generated data. To the best of our knowledge, this is the first paper that thoroughly studies the possibility of generating a rich collection of time series. Our method not only generates realistic time series data but also gives a higher coverage of the feature space than existing time series benchmarking data.

The GRATIS approach is also able to efficiently generate new time series with controllable target features, by tuning the parameters of MAR models. This is particularly useful in time series classification or specific areas where only some features are of interest. This procedure is the inverse of feature extraction which usually requires much computational power. Our approach of generating new time series from given features can scale up the computation time by 40 times (compared to Kang et al., 2017) making feature-driven time series analysis tasks feasible.

We further show that the GRATIS scheme can serve as a useful resource for time series applications. In particular, we present a novel time series forecasting approach by exploiting the time series features of current generated time series. Our application also sheds light on a potential direction to forecasting with private data where the model training could be purely based on our generated data. Other potential extensions include: (i) GRATIS with exogenous information via mixture of ARIMA with explanatory variables (ARIMAX) to allow for local patterns due to external events, (ii) GRATIS with multivariate time series by exploring mixtures of vector autoregression models, (iii) GRATIS with cross-sectional information about the time series by exploring the approaches in Wang et al. (2008) and Ashouri et al. (2019), (iv) extending GRATIS to discrete time series by investigating the mixture of integer-valued autoregressive processes Latour, 1998 or Poisson autoregression (Freeland and McCabe, 2004), and (v) using GRATIS to serve as a pre-training process of deep learning methods to save time and improve accuracy.

# 7 Acknowledgments

The authors are grateful to Associate Editor, two anonymous reviewers for their helpful comments that improved the contents of this paper. Yanfei Kang and Feng Li are supported by the National Natural Science Foundation of China (No. 11701022 and No. 11501587, respectively). Rob J Hyndman is supported by the Australian Centre of Excellence in Mathematical and Statistical Frontiers.

# A  Appendix

## A.1  Description of time series features

In this section, we document the feature details in Table 3. We have also developed an R package `tsfeatures` to provide methods for extracting various features from time series data which is available at https://CRAN.R-project.org/package=tsfeatures.. Note that all of our features are ergodic for stationary and difference-stationary processes, and not dependent on the scale of the time series. Thus, they are well-suited for applying to a large diverse set of time series.

Each time series, of any length, can be summarized as a feature vector $\boldsymbol{F} = (F_1, F_2, \ldots, F_{26})'$. The length of this vector will be 42 for non-seasonal time series, and for seasonal time series with a single seasonal period. For multiple seasonal time series, $\boldsymbol{F}$ will have a few more elements.

The first five features in Table 3 are all positive integers. $F_1$ is the time series length. $F_2$ is the number of seasonal periods in the data (determined by the frequency of observation, not the observations themselves) and set to 1 for non-seasonal data. $F_3$ is a vector of seasonal periods and set to 1 for non-seasonal data. $F_4$: the number of first-order differences required before the data pass a KPSS stationarity test (Kwiatkowski et al., 1992) at the 5% level. $F_5$ is the number of seasonal differences required before the data pass an OCSB test (Osborn et al., 1988) at the 5% level. For multiple seasonal time series, we compute $F_5$ using the largest seasonal period. For non-seasonal time series (when $F_1 = 1$), we set $F_5 = 0$.

We compute the autocorrelation function of the series, the differenced series, and the twice-differenced series. Then $F_6$ is a vector comprising the first autocorrelation coefficient in each case, and the sum of squares of the first 10 autocorrelation coefficients in each case. The autocorrelation coefficient of the original series at the first seasonal lag is also computed. For non-seasonal data, this is set to 0.

We compute the partial autocorrelation function of the series, the differenced series, and the second-order differenced series. Then $F_7$ is a vector comprising the sum of squares of the first 5 partial autocorrelation coefficients in each case. The partial autocorrelation coefficient of the original series at the first seasonal lag is also computed. For non-seasonal data, this is set to 0.

The spectral entropy is the Shannon entropy

$$F_8 = -\int_{-\pi}^{\pi} \hat{f}(\lambda) \log \hat{f}(\lambda) d\lambda,$$

where $\hat{f}(\lambda)$ is an estimate of the spectral density of the data. This measures the "forecastability" of a time series, where low values of $F_8$ indicate a high signal-to-noise ratio, and large values of $F_8$ occur when a series is difficult to forecast.

The nonlinearity coefficient ($F_9$) is computed using a modification of the statistic used in Teräsvirta's nonlinearity test. Teräsvirta's test uses a statistic $X^2 = T \log(\text{SSE1}/\text{SSE0})$ where SSE1 and SSE0 are the sum of squared residuals from a nonlinear and linear autoregression respectively. This is non-ergodic, so is unsuitable for our purposes. Instead, we define $F_9 = X^2/T$ which will converge to a value indicating the extent of nonlinearity as $T \to \infty$.

We use a measure of the long-term memory of a time series ($F_{10}$), computed as 0.5 plus the maximum likelihood estimate of the fractional differencing order $d$ given by Haslett and Raftery (1989). We add 0.5 to make it consistent with the Hurst coefficient. Note that the fractal dimension can be estimated as $D = 2 - F_9$.

$F_{11}$ and $F_{12}$ are two time series features based on tiled (non-overlapping) windows. Means or variances are produced for all tiled windows. Then stability is the variance of the means, while lumpiness is the variance of the variances.

$F_{13}$ is a vector comprising the statistic for the KPSS unit root test with linear trend and lag one, and the statistic for the "Z-alpha" version of PP unit root test with constant trend and lag one.

The next three features ($F_{14}$, $F_{15}$, $F_{16}$) compute features of a time series based on sliding (overlapping) windows. $F_{14}$ finds the largest mean shift between two consecutive windows. $F_{15}$ finds the largest variance shift between two consecutive windows. $F_{16}$ finds the largest shift in Kulback-Leibler divergence between two consecutive windows.

The following six features ($F_{17}$– $F_{22}$) are modifications of features used in Kang et al. (2017). We extend the STL decomposition approach (Cleveland et al., 1990) to handle multiple seasonalities. Thus, the decomposition contains a trend, up to $M$ seasonal components, and a remainder component:

$$x_t = f_t + s_{1,t} + \cdots + s_{M,t} + e_t$$

where $f_t$ is the smoothed trend component, $s_{i,t}$ is the $i$th seasonal component and $e_t$ is a remainder component. The components are estimated iteratively. Let $s_{i,t}^{(k)}$ be the estimate of $s_{i,t}$ at the $k$th iteration, with initial values given as $s_{i,t}^{(0)} = 0$. Then we apply an STL decomposition to $x_t - \sum_{j \neq i}^{j=1\,M} s_{j,t}^{(k-1)}$ to obtained updated estimates $s_{i,t}^{(k)}$ for $k = 1, 2, \ldots$. In practice, this converges quickly and only two iterations are required. To allow the procedure to be applied automatically, we set the seasonal window span for STL to be 21 in all cases. For a non-seasonal time series (when $F_1 = 1$), we simply estimate $x_t = f_t + e_t$ where $f_t$ is computed using Friedman's "super smoother" (Friedman, 1984).

- $F_{17}$ and $F_{18}$ are defined as

$$F_{17} = 1 - \frac{\text{Var}(e_t)}{\text{Var}(f_t + e_t)} \qquad \text{and} \qquad F_{18,i} = 1 - \frac{\text{Var}(e_t)}{\text{Var}(s_{i,t} + e_t)}.$$

  If their values are less than 0, they are set to 0, while values greater than 1 are set to 1. For non-seasonal time series $F_{18} = 0$. For seasonal time series, $F_{18}$ is an $M$-vector, where $M$ is the number of periods. This is analogous to the way the strength of trend and seasonality were defined in Wang et al. (2006), Hyndman et al. (2015) and Kang et al. (2017).

- $F_{19}$ measures the "spikiness" of a time series, and is computed as the variance of the leave-one-out variances of the remainder component $e_t$.

- $F_{20}$ and $F_{21}$ measures the linearity and curvature of a time series calculated based on the coefficients of an orthogonal quadratic regression.

- We compute the autocorrelation function of $e_t$, and $F_{22}$ is a 2-vector containing the first autocorrelation coefficient and the sum of the first ten squared autocorrelation coefficients.

The remaining features measure the heterogeneity of the time series. First, we pre-whiten the time series to remove the mean, trend, and autoregressive (AR) information (Barbour and Parker, 2014). Then we fit a GARCH(1,1) model to the pre-whitened time series, $x_t$, to measure for autoregressive conditional heteroskedasticity (ARCH) effects. The residuals from this model, $z_t$, are also measured for ARCH effects using a second GARCH(1,1) model.

- $F_{23}$ is the sum of squares of the first 12 autocorrelations of $\{x_t^2\}$.

- $F_{24}$ is the sum of squares of the first 12 autocorrelations of $\{z_t^2\}$.

- $F_{25}$ is the $R^2$ value of an AR model applied to $\{x_t^2\}$.

- $F_{26}$ is the $R^2$ value of an AR model applied to $\{z_t^2\}$.

The statistics obtained from $\{x_t^2\}$ are the ARCH effects, while those from $\{z_t^2\}$ are the GARCH effects. Note that the two $R^2$ values are used in the Lagrange-multiplier test of Engle (1982), and the sum of squared autocorrelations are used in the Ljung-Box test proposed by Ljung and Box (1978).

## A.2 Web application for time series generation

We implement our approach in a shiny app (Chang et al., 2018) as shown in Figure 9. Users can first choose the features that they are interested in. After setting their desired time series seasonal pattern, length, the number of time series, and the feature values required, the generated series are displayed. In Figure 9, we aim to generate ten monthly time series with length 120. The selected features are nsdiffs, x_acf1, entropy, stability, trend, seasonal_strength and garch_r2. The corresponding target vector is $(1, 0.85, 0.55, 0.73, 0.91, 0.95, 0.07)'$. Following the GA process, the generated time series are shown at the bottom of Figure 9. The simulated data can also be downloaded to a local computer.

## A.3 Penalized linear quantile regression

Consider a sample $\{y_i, x_i\}$ for $i = 1, ..., n$ where $y_i$ is the MASE values and $x_i$ is the feature vector in our framework. The conditional $\tau$th quantile function is defined as $f_\tau(x_i)$ where $\tau = p(y_i < f_\tau(X_i)|X_i = x_i)$. Following Koenker and Bassett Jr (1978), we define the check function
$$\rho_\tau(r) = \begin{cases} \tau r & r > 0; \\ -(1-\tau)r & \text{otherwise.} \end{cases}$$

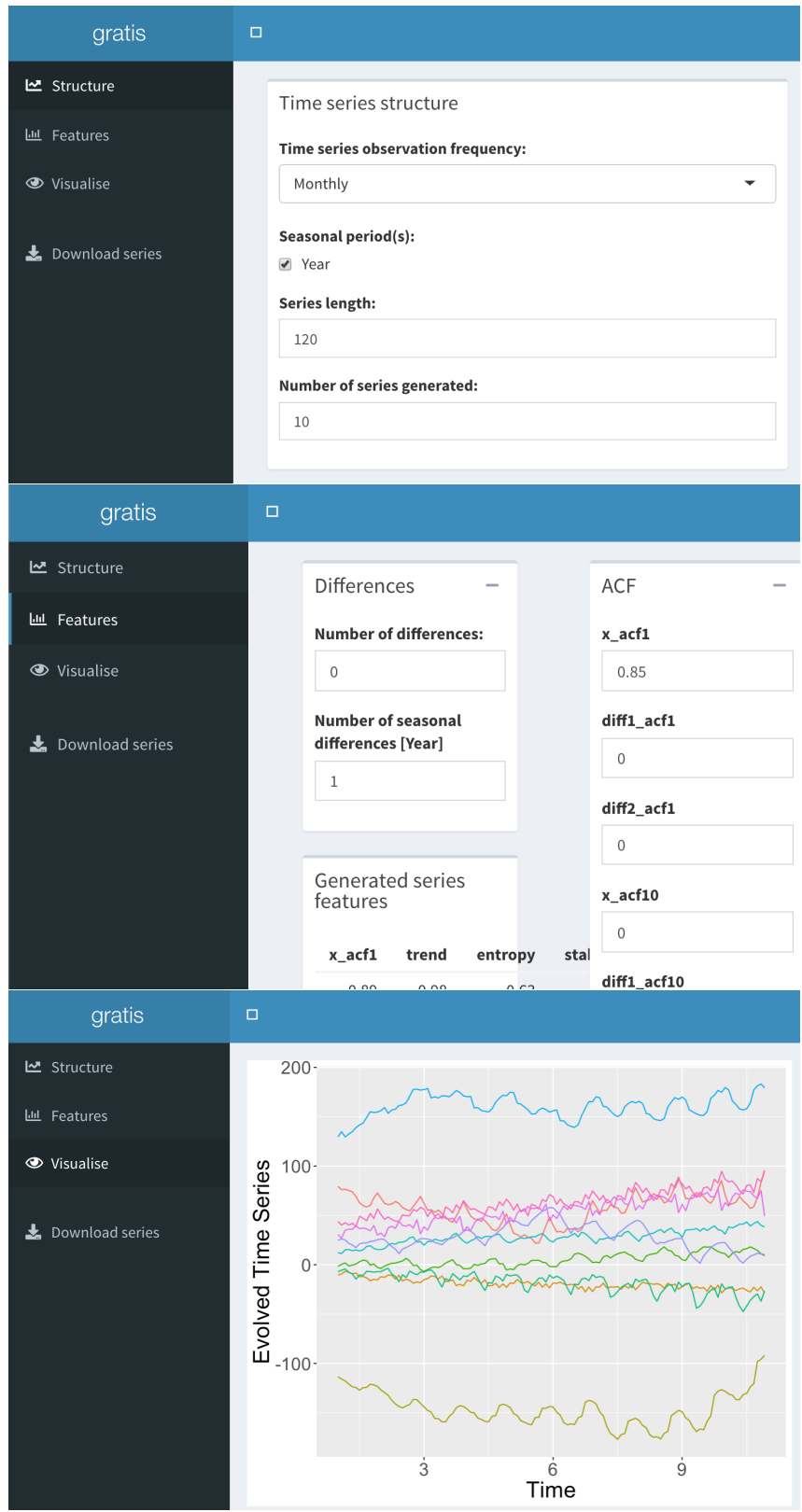

Figure 9: The shiny application generating time series with controllable features. The application is designed with four tabs, named 'Structure', 'Features', 'Visualise' and 'Download series'. The top panel shows the options for the basic structure of the desired time series. For illustration, this interface illustrates the generation of ten monthly time series with length 120. The middle panel shows specification of the target feature vector, which consists of the non-zero values set for the features. The generated ten time series with the target features are shown at the bottom panel. The app is available at `https://ebsmonash.shinyapps.io/tsgeneration/`.

Thus the $\tau$th conditional quantile regression with the adaptive LASSO penalty can be estimated by solving the following minimization problem

$$\min \sum_{i=1}^{n} \rho_\tau(y_i - x_i'\beta) + \lambda \sum_{j=1}^{d} \omega_j |\beta_j|$$

for some appropriately chosen $\lambda$ and weights $\omega_j$.

With the aid of slack variables, the convex optimization problem can be equivalently written into a linear programming problem which can be easily solved by standard optimization software. A comparison of existing algorithms can be found in Peng and Wang (2015).

## A.4 Genetic Algorithms

Genetic algorithms are stochastic search algorithms which are able to solve optimization problems. GAs use evolutionary strategies inspired by the principles of biological evolution where parameters are treated as population. At a certain stage of evolution a population is composed of a number of chromosomes. Applying some mapping from the chromosome representation into the decision variable space, which represents potential solutions to an optimization problem.

A typical genetic algorithm requires three inputs, (i) type of variables (ii) fitness function, and (iii) convergence criteria which is usually the improvement of fitness function. Algorithm 1 documents the evolution process of the genetic algorithms.

---
**Algorithm 1:** Genetic algorithms

**Output:** Evolved population
**Input :** Initial population; Fitness function; Convergence criteria
**while** *the convergence criteria not met* **do**
- Run the genetic operators

  1. **Selection**: mimics the behavior of natural organisms in a competitive environment, in which only the most qualified and their offspring survive. It samples the population with replacement where each individual has a given probability of surviving.
  2. Form the new population
     (a) **Crossover**: forms new offsprings from two parent chromosomes by combining part of the genetic information from each.
     (b) **Mutation**: randomly alters the values of genes in a parent chromosome.

- Fitness evaluation with the fitness function.

**end**

---